\documentclass[letterpaper,10pt,conference]{ieeeconf}

\IEEEoverridecommandlockouts
\usepackage{graphicx} 
\usepackage{subcaption}
\usepackage{amsmath} 
\usepackage{amssymb} 
\usepackage{bm}
\usepackage{hyperref}

\usepackage{amsmath}
\DeclareMathOperator*{\argmax}{arg\,max}

\usepackage{cleveref}
\crefname{table}{Table}{Tables}
\crefname{figure}{Figure}{Figures}
\crefname{section}{Section}{Sections}

\usepackage[binary-units]{siunitx}
\usepackage[acronym]{glossaries}
\newacronym{mse}{MSE}{mean squared error}
\newacronym{mlp}{MLP}{Multi-Layer Perceptron}
\newacronym{fmcw}{FMCW}{Frequency-Modulated Continuous-Wave}
\newacronym{mmw}{mm-W}{millimetre-Wave}
\newacronym{nn}{NN}{Neural Network}
\newacronym{ro}{RO}{Radar Odometry}
\newacronym{aspp}{ASPP}{Atrous Spatial Pyramid Pooling}
\newacronym{slam}{SLAM}{Simulaneous Localisation And Mapping}
\newacronym{rrcd}{RRCD}{Radar RobotCar Dataset}
\newacronym{rdd}{RDD}{Radar Doppler Dataset}

\title{\bf Doppler-aware Odometry from FMCW Scanning Radar}
\author{Fraser Rennie, David Williams, Paul Newman and Daniele De Martini
\\
Oxford Robotics Institute, Dept. Engineering Science, University of Oxford, UK.\\\texttt{fraser.rennie@exeter.ox.ac.uk}, \texttt{\{dw,pnewman,daniele\}@robots.ox.ac.uk}
\\
\thanks{This project was supported by EPSRC Programme Grant ``From Sensing to Collaboration'' (EP/V000748/1). We also thank the teams from Oxbotica and Navtech Radar for their invaluable help in collecting the dataset and support on this work.}
}

\usepackage{copyright}

\begin{document}

\maketitle
\copyrightnotice

\begin{abstract}
    This work explores Doppler information from a \gls{mmw} \gls{fmcw} scanning radar to make odometry estimation more robust and accurate.
    Firstly, doppler information is added to the scan masking process to enhance correlative scan matching. 
    Secondly, we train a \gls{nn} for regressing forward velocity directly from \textit{a single} radar scan; we fuse this estimate with the correlative scan matching estimate and show improved robustness to bad estimates caused by challenging environment geometries, e.g. narrow tunnels.
    We test our method with a novel custom dataset which is released with this work at \url{https://ori.ox.ac.uk/publications/datasets}. 
\end{abstract}


\glsresetall

\section{Introduction}
Autonomous vehicles start to play a more important and widespread role in transportation as more and more countries are amending road regulations to allow their deployment\footnote{https://www.gov.uk/government/news/self-driving-revolution-to-boost-economy-and-improve-road-safety}.

\begin{figure}[h!]
    \centering
    \includegraphics[angle=0,width=0.28\textwidth]{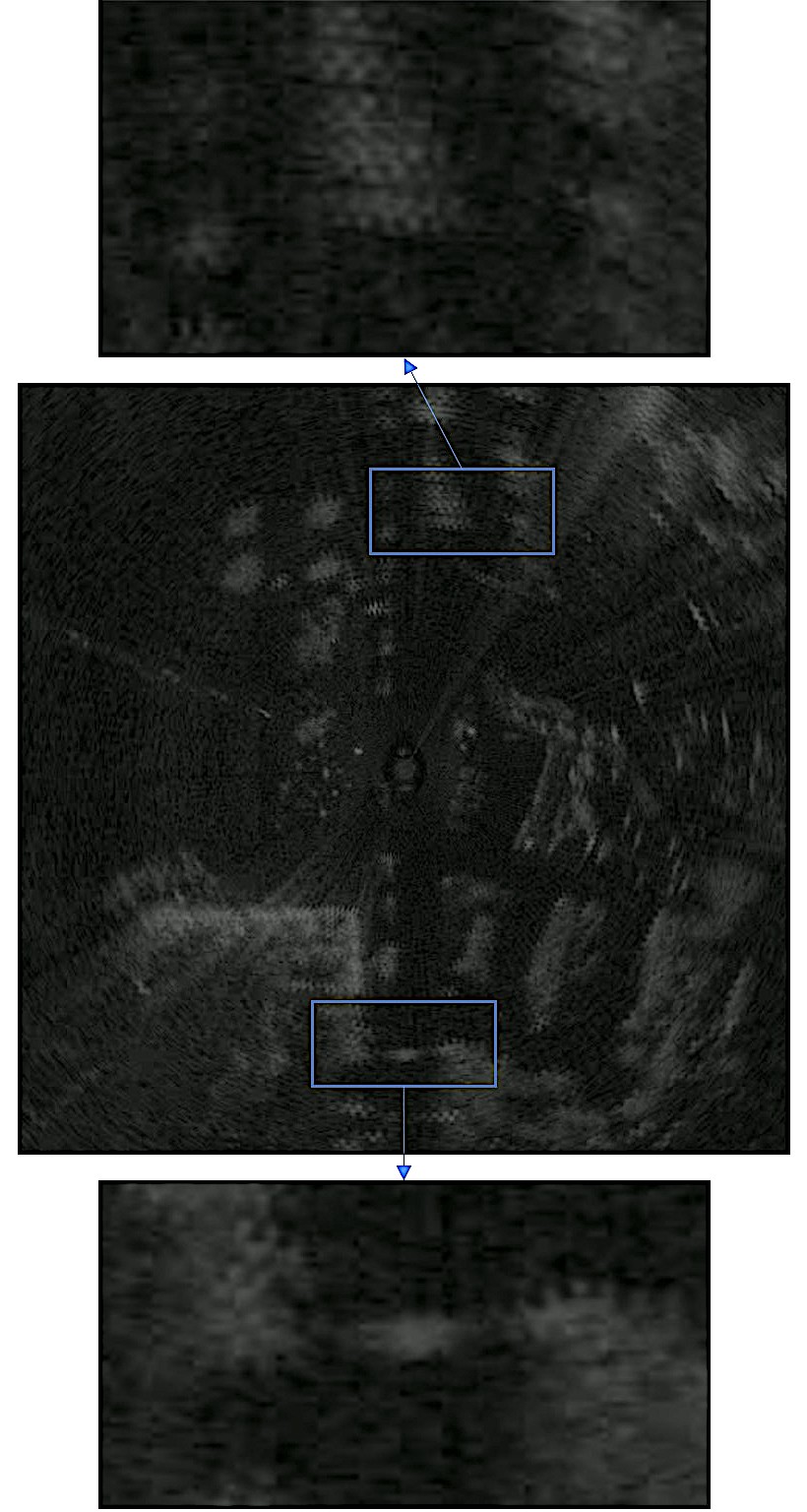}
    \caption{Radar scan from the RDD dataset. In the right portion, two regions extracted show the ``zig-zag'' pattern caused by the alternating modulation patterns -- in conjunction with the ego-vehicle speed. The top depicts a static object which is affected uniformly by it, while, at the bottom, a chase car with the same speed as the ego-vehicle is unaffected.\label{fig:zig_zag}}
    \vspace{-15pt}
\end{figure}

Reliable ego-motion estimation and localisation are fundamental components of autonomous navigation, answering the question of where the vehicle is located and how it moves in the environment.
As considered deployment scenarios become more challenging, the detection methods and the sensors collecting data about a vehicle's surroundings must advance.
Currently, the primary sensors used by autonomous vehicles are cameras and LiDAR: while these traditional sensors may perform adequately under favourable conditions, their effectiveness can be significantly compromised by factors such as rain, snow, fog and low light.
In contrast, radar offers distinct advantages, including a superior range and invariance to adverse weather and low light conditions \cite{burnett2023boreas}.

We explore exploiting Doppler information in radar scans for odometry.
We first evaluate its utility for masking distractors in radar scans before motion estimation through cross-correlation.
We then rely solely on Doppler for motion estimation by training a regressor directly on single scans. 

Although Doppler information is usually delivered from radar sensors together with intensity, no available datasets provide it for \gls{fmcw} scanning radars \cite{burnett2023boreas,9197298,barnes2020oxford,sheeny2020radiate}.
Thus, we collected a custom dataset -- an example image is shown in \cref{fig:zig_zag} -- which we used to train and evaluate our approach.
On such a dataset, we demonstrate how Doppler information can improve the masking approach \textit{and} generate reliable forward-motion estimation when used for direct regression, even in feature-poor scenery such as narrow tunnels.


To summarise, the main contributions of this work are:
\begin{itemize}
    \item Use of Doppler information for the masking process;
    \item Single-scan longitudinal relative-pose estimation;
    \item A novel \gls{fmcw} radar dataset containing Doppler information, released alongside this work.
\end{itemize}

\section{Related work}

\Gls{fmcw} scanning radar has seen an increasing interest in the past years, from odometry and localisation \cite{gadd2021contrastive,de2020kradar,suaftescu2020kidnapped,9197298} to semantic segmentation \cite{kaul2020rss,8794014}, detection \cite{sheeny2020radiate}, path planning \cite{broome2020road,williams2020keep} and cross-modal localisation \cite{tang2021self,tangauro2023}.
As this work discusses odometry, we will focus on this topic.
Odometry can be classified into two main groups, direct and indirect: the first uses direct data collected from the sensors whereas the second requires preprocessing, usually in the form of landmark detection.
Both families of algorithms have been explored for \gls{fmcw} scanning \gls{ro}.

The works \cite{8460687,cen2019radar} demonstrated the potential of \gls{fmcw} scanning radar for odometry.
Such indirect-odometry methods focus on resilient landmark extraction -- based on shape similarity and a gradient-based feature detector respectively -- and a robust segment-based matching algorithm to deliver \gls{ro} competitive with vision and LiDAR.
\cite{8794014} applied a learnt method to reject landmarks appearing on unwanted distractor artefacts, speeding up the sequent scan-matching process.

Other \gls{ro} indirect methods include Cell Averaging CFAR (CA-CFAR) \cite{9100374}, Bounded False Alarm Rate (BFAR) \cite{alhashimi2021bfarbounded} and Conservative Filtering for Efficient and Accurate Radar Odometry (CFEAR) \cite{cfear}; these methods focus on improving the feature extraction through hand-crafted landmark detection algorithms.
Other methods learn feature detectors from the data: \cite{barnes2020under} uses a U-Net network to predict a map of landmark locations, scores and descriptions from odometry annotations.
In \cite{burnett_radar_odom_2021}, the learning approach is pushed even further by training the network unsupervised, jointly estimating landmarks and relative poses.

Differently, direct methods estimate odometry directly from the raw scan.
Our method belongs to this family; in particular, we base our scan-matching procedure on \cite{barnes2020masking,weston2022fastmbym}.
\cite{barnes2020masking} employs a \gls{nn} to generate a mask then used to filter out distractors from the radar scans.
The relative pose is then estimated using correlative scan matching and, through supervision from odometry directly, the masking network is trained.
\cite{weston2022fastmbym} improves this method's time and memory performances by exploiting the Fourier Transform's properties.
Our approach applies the same technique as \cite{weston2022fastmbym}, where we tailor the network to use Doppler information as an additional source.

Doppler effect has been explored for \gls{fmcw} scanning radars in \cite{9327473}, where the authors evaluated its effect on radar odometry.
Here, the authors propose a compensation method to correct return displacements based on their estimated velocity.
As the data did not contain multiple modulations or direct Doppler measurement, the speed needed to be estimated from multiple scans and the effect of the correction on the final accuracy was negligible.

Doppler information has also been exploited in the literature to solve odometry.
For instance, \cite{zhuang20234d} exploits landmark velocities to assist a radar-based \gls{slam} pipeline; yet the estimates are given as a pointcloud, and the raw data is not accessed directly.
Doppler has also been explored in a recent work on \gls{fmcw} LiDAR \cite{wu2022picking}, where the Doppler velocity measurements are used to assist relative pose estimations.


\section{Preliminaries and dataset}

The sensor we employ is the Navtech CTS350-X, a \gls{fmcw} scanning radar.
The sensor spins at \SI{4}{\hertz}, transmitting a continuous radio wave; however, the radio wave frequency is modulated, i.e. changed throughout the operation, allowing for object-distance measurements by comparing the frequency of the transmitted and returned waves.

\begin{figure}
    \centering
    \begin{subfigure}{0.45\textwidth}
    \includegraphics[trim=0 0 0 20,clip,width=0.9\textwidth]{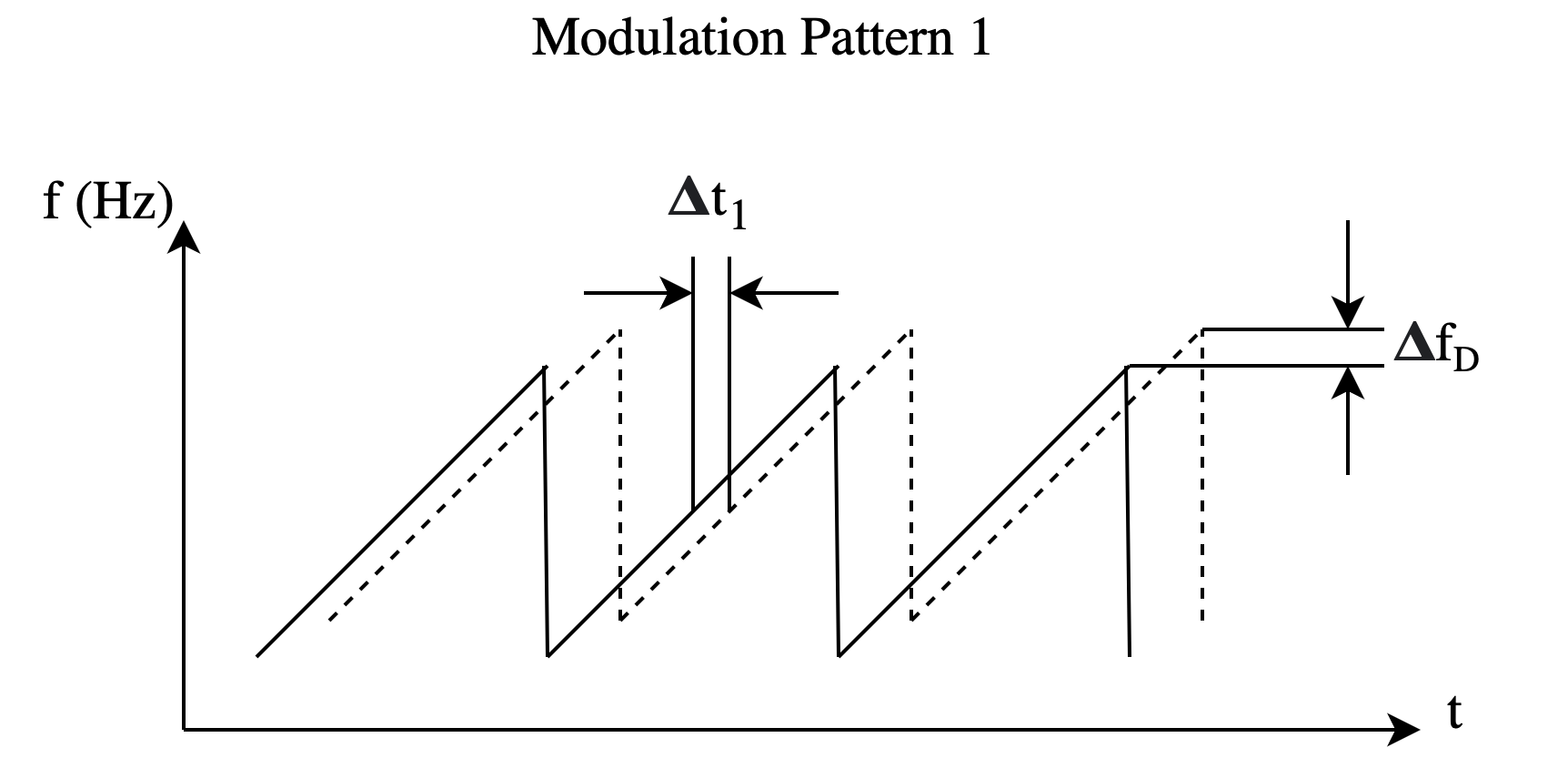}
    \caption{}
    \label{fig:modulation_pattern_1}
    \end{subfigure}
    
    \begin{subfigure}{0.45\textwidth}
    \includegraphics[trim=0 0 0 20,clip,width=0.9\textwidth]{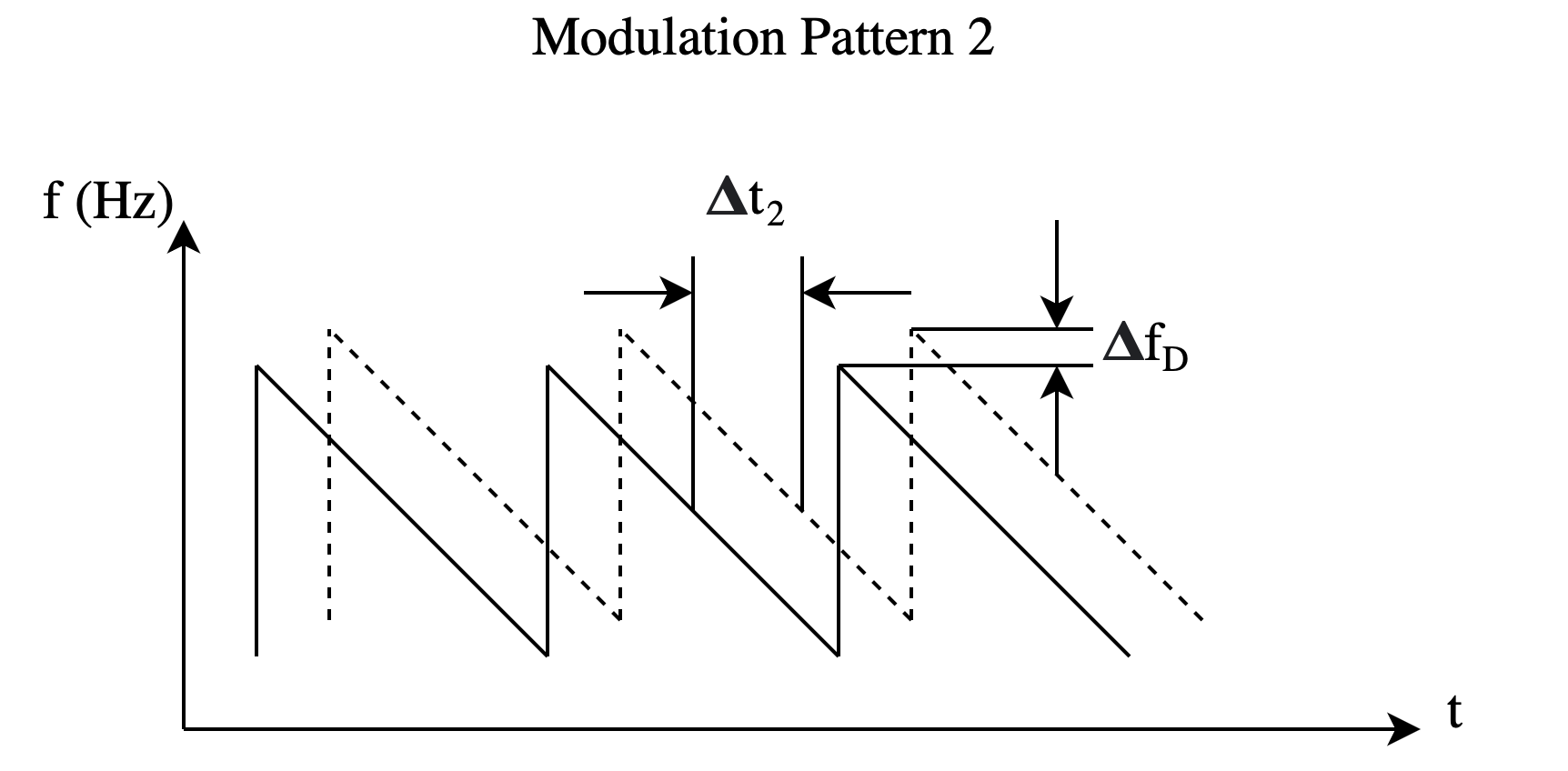}
    \caption{}
    \label{fig:modulation_pattern_2}
    \end{subfigure}
    \caption{FMCW Modulation patterns: (a) a transmitted sawtooth wave (solid line) is reflected and distorted by the object to produce the reflection wave (dotted line) used to estimate the distance of the reading; (b) a reverse sawtooth wave transmitted onto the same objects can be compared to (a) to extract Doppler information.\label{fig:modulations}}
    \vspace{-15pt}
\end{figure}

In this system, the echo signal’s frequency has a delay of $\Delta t$ called a \textit{runtime shift}, calculated via the change in frequency, also called the \emph{beat frequency}.
This is shown in \cref{fig:modulation_pattern_1}, where $\Delta f_1$ is the beat frequency.

The distance $R$ to a given object is then given by:
\begin{equation}
    R=\frac{c_o|\Delta t|}{2}= \frac{c_0|\Delta f|}{2\big(\frac{\text d f}{\text d t}\big)}
    \label{eqn:range_radar}
\end{equation}
where $\frac{\text df}{\text dt}$ is the gradient of the sawtooth wave.
Dividing the beat frequency by the gradient gives the runtime shift, $\Delta t$, which then allows for a direct calculation of the range of the object from which the echo is transmitted.

\subsection{Doppler Effect}

If an object is non-stationary, the term $\Delta f$ becomes influenced by the object's speed towards the radar sensor, affecting the perceived range estimation due to the Doppler effect introduced into the system.
The beat frequency, $\Delta f$, will then contain a frequency shift due to the target’s range, runtime shift $\Delta t$, and a Doppler frequency shift, $\Delta f_D$, which is shown as an error in the range reading.

Let's now gather a radar reading of the same azimuth with a different modulation pattern, as in \cref{fig:modulation_pattern_2}.
This new modulation pattern allows for a comparison between range errors as the received waves have been influenced in opposite directions due to the opposite sawtooth gradients.

The Doppler frequency is calculated via:
\begin{equation}
    \Delta f = \frac{\Delta v}{c}f_e
\end{equation}
where $\Delta v$ is the relative radial velocity between the vehicle and object, $c$ is the light speed and $f_e$ the emitted frequency.

While rotating, the radar sensor inspects one angular portion (azimuth, $\alpha$) of space at a time.
A complete scan is composed of $N_\alpha = 400$ azimuths. 
The sensor has been set up to utilise the two sawtooth waves alternately, standard for odd and reversed for even azimuths.
This approach approximates gathering the same environment with the two modalities, as the azimuths overlap.
This results in polar radar images of shape $N_\alpha \times N_b = 400 \times 3600$, where $N_b$ is the number of bins, each with a size of \SI{4.38}{\centi\metre}, which can be transformed in cartesian coordinates, as shown in \cref{fig:doppler_normal_route_scans}.

\subsection{Radar Doppler Dataset (RDD)}

No available datasets include scanning \gls{fmcw} radar with Doppler information.
For this reason, we propose the \gls{rdd}, a novel dataset containing both structured and unstructured environments to stretch-test the contribution of Doppler information to motion estimation.

\Cref{fig:zig_zag} depicts one image from the proposed dataset.
Notably, the car is moving, and a chase car follows it at roughly the same speed.
Clear ``zigzag'' artefacts directly result from the two different perceived distances of the objects: the environment is affected by the artefacts depending on the angle of perception, as the Doppler shift depends on the direction of movement.
The chase car, instead, is practically unaffected as its speed matches the ego vehicle.

\begin{figure}
\centering
\begin{subfigure}{.37\textwidth}
  \centering
  \includegraphics[width=1.0\textwidth]{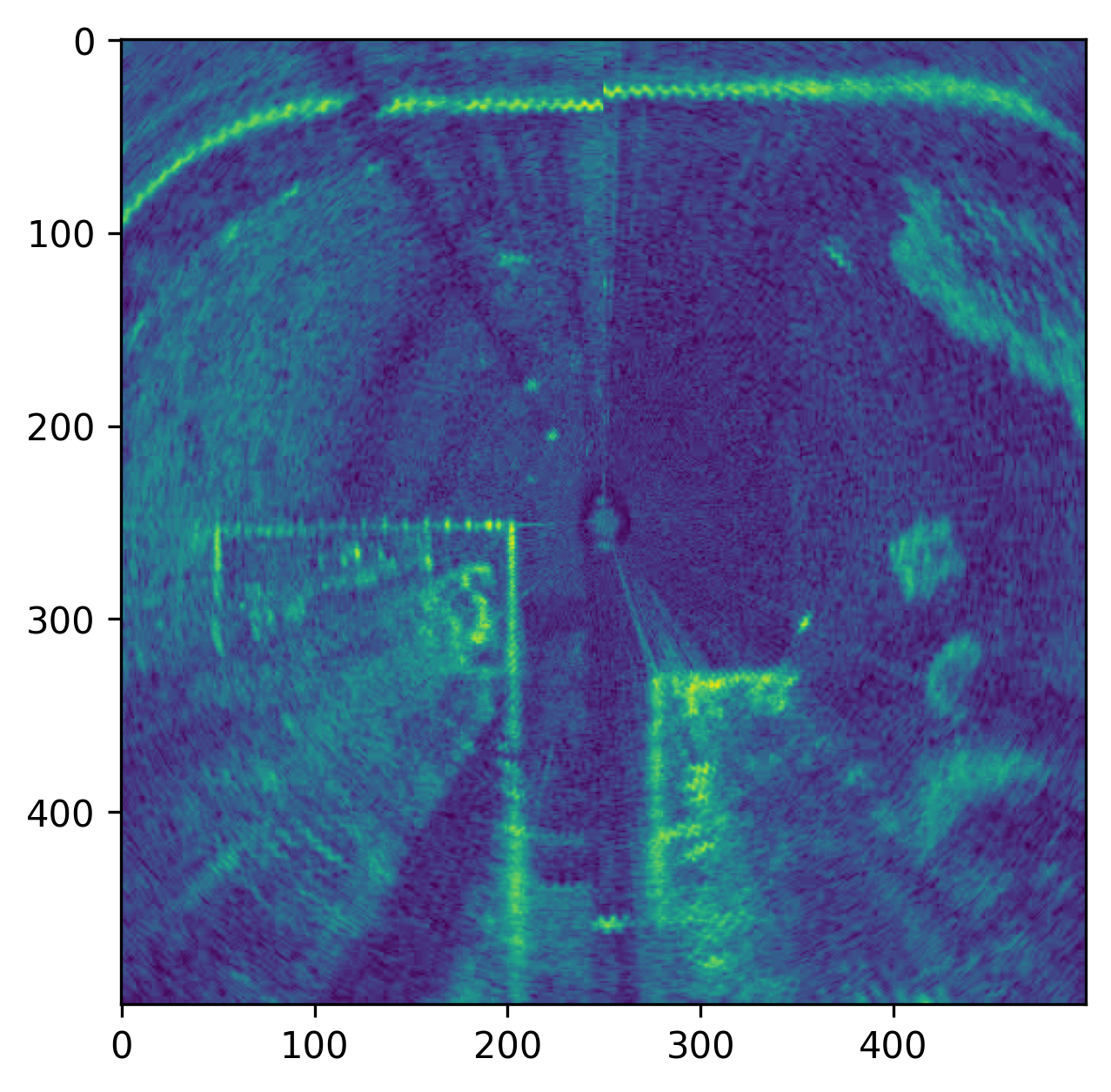}
  \caption{\label{fig:doppler_normal_val_scan}}
\end{subfigure}

\begin{subfigure}{.37\textwidth}
  \centering
  \includegraphics[width=1.0\textwidth]{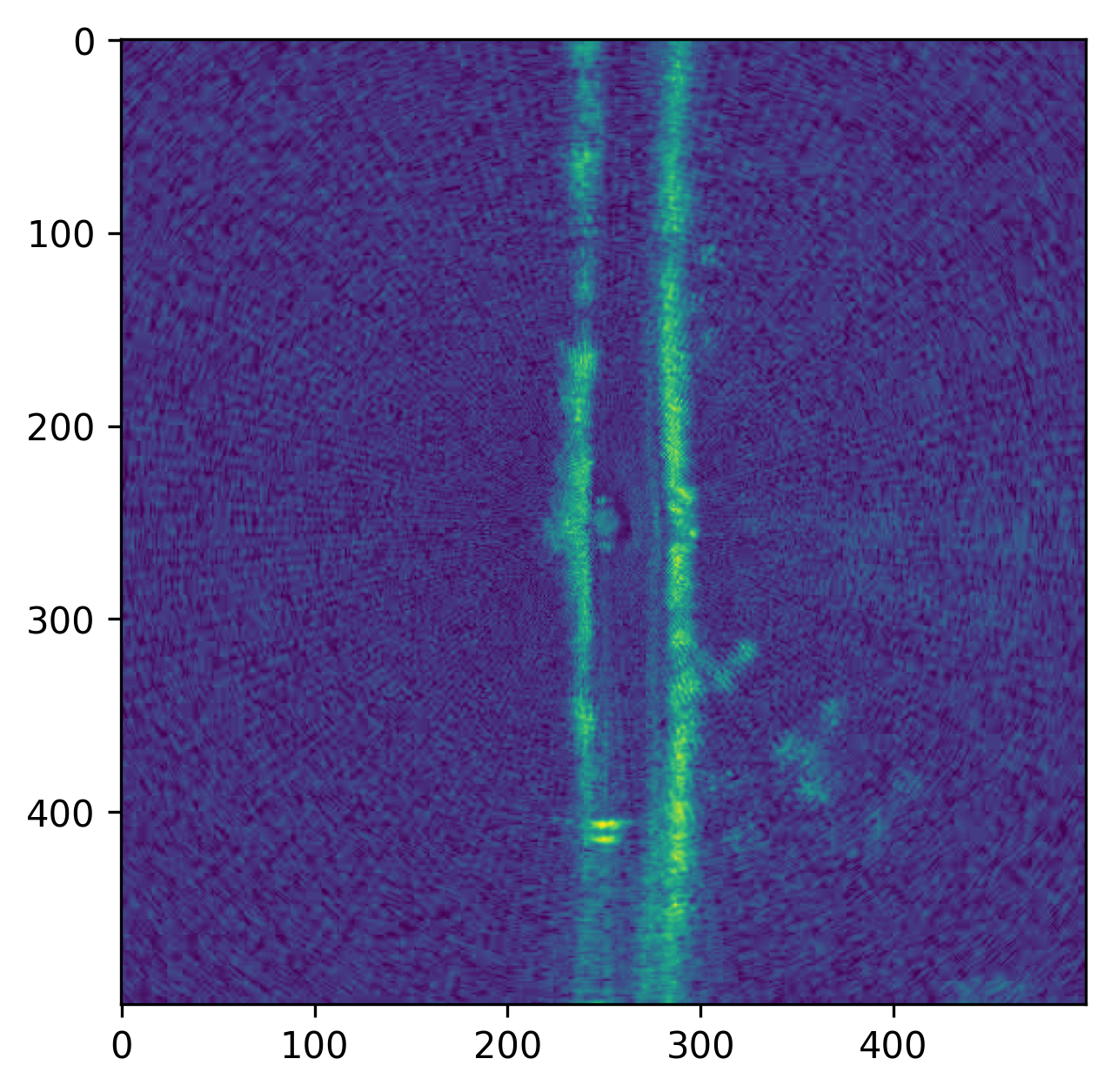}
  \caption{\label{fig:doppler_normal_test_scan}}
\end{subfigure}

\caption{Radar scans in the \gls{rdd} show (a) structured and (b) unstructured scenes.}
\label{fig:doppler_normal_route_scans}
\vspace{-13pt}
\end{figure}

Various routes were collected to maximise the coverage of Doppler effect artefacts, including driving:
\begin{itemize}
    \item at \SI{10}{mph} towards and away from a wall;
    \item along a \SI{4.5}{\kilo\meter} stretch of a geometrically degraded road;
    \item around the feature-rich Culham Science Centre.
\end{itemize}
The total driving amounts to \SI{24.7}{\kilo\meter}, for a duration of \SI{59}{\minute}.
\Cref{fig:doppler_normal_route_scans} shows examples of feature-poor and feature-rich scans.
In particular, \cref{fig:doppler_normal_test_scan} shows a very challenging scene for standard \gls{ro} pipelines, as the narrow tunnel is practically unaffected visibly by the vehicle's movement, and so odometry is likely to fail.

The dataset is released alongside other sensor readings, especially cameras and LiDARs.
Importantly, RTK GPS is used to create the ground-truth annotations exploited in this work to learn to estimate odometry.

\section{Methodology}
Here, we introduce the correlative scan-matching approach to retrieve the ego-motion between two radar scans.
We then describe how we integrate Doppler information and fuse such estimate with a pure Doppler-based velocity estimation.

\begin{figure*}
    \centering
    \begin{subfigure}{0.83\textwidth}
    \centering
        \includegraphics[width=0.9\textwidth]{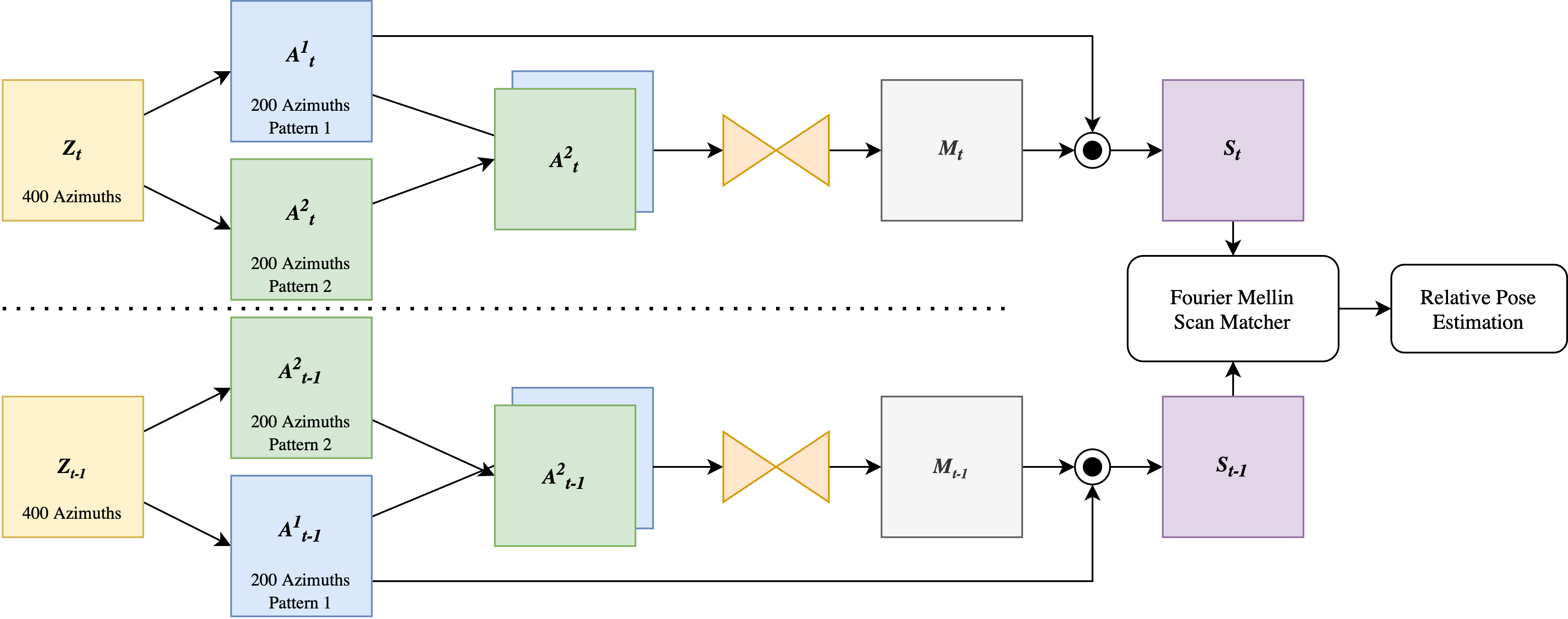}
        \caption{\label{fig:2_channel_diagram}}
    \end{subfigure}
    \vspace{5pt}
    
    \begin{subfigure}{0.77\textwidth}
    \centering
        \includegraphics[width=0.85\textwidth]{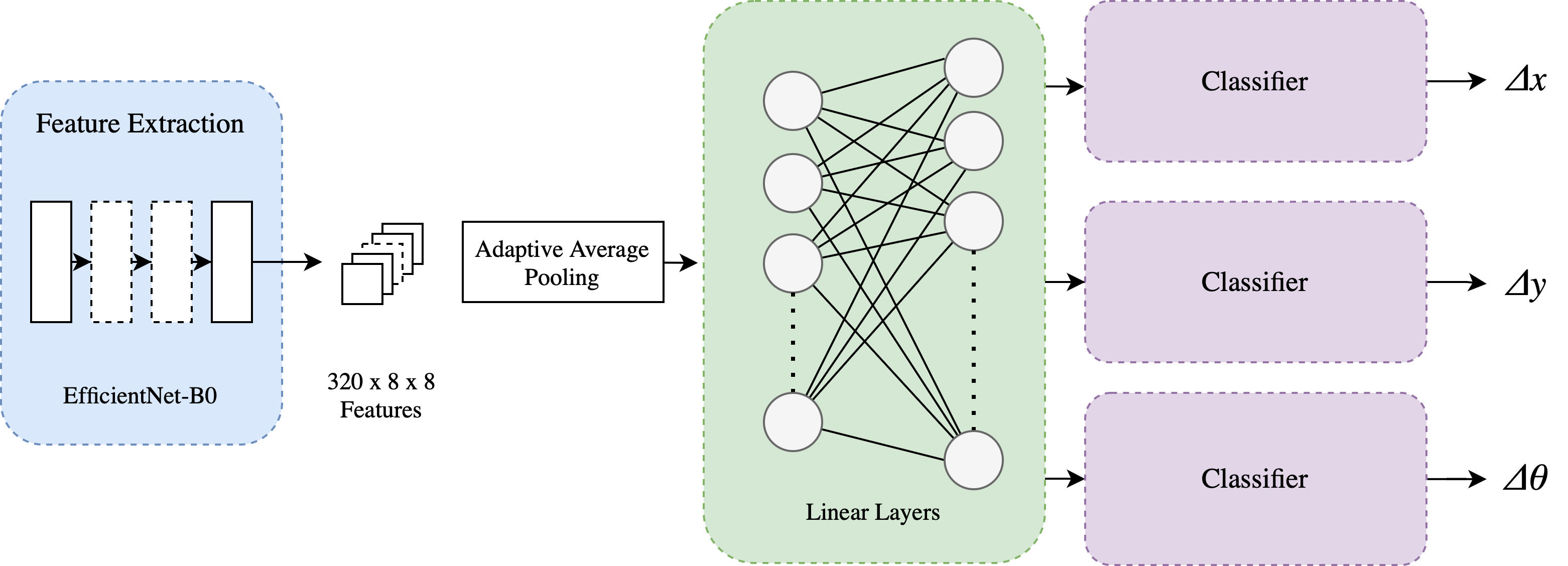}
        \caption{\label{fig:novel_network_diagram}}
    \end{subfigure}
    \caption{(a) Our Doppler-aware scan matching generates a mask from the 2-channel images of the two modulations for both time steps.
    It is then applied to the modulation-patter-1 scans via element-wise product. The resultant scans are used in the scan-matching process. (b) DAVE-NN is composed of a feature extractor and three regression heads. These are implemented to solve regression through an intermediate classification step.\label{fig:method}}
    \vspace{-10pt}
\end{figure*}

\subsection{Correlative scan-matching}

Let $Z_{t-1}$ and $Z_t$ be two consecutive radar scans in cartesian form in $\mathbb{R}^{n\times n}$, where $n$ is the dimension in pixels of the images.
Our goal is to retrieve a rigid-body transformation $[R | t] \in \mathbb{SO}(2)$ -- where $R$ is a rotation of an angle $\theta$ and $t$ a translation $[t_x, t_y]$ -- such as $Z_t \approx T_{[R|t]}(Z_{t-1})$, where $T_{[R|t]}$ is an affine transformation following $R$ and $t$.
Such an expression is approximate as the radar signals suffer appearance change due to noises and artefacts.

A correlative scan-matching approach can find the optimum $[R^*|t^*]$ as the transformation that maximises the correlation between the two signals $Z_{t-1}$ and $Z_t$, such as:
\begin{equation}
\label{eq:correlatoin}
    [R^*|t^*] = \argmax_{[R|t] \in \mathbb{SO}(2)} T_{[R|t]}(Z_{t-1}) \star Z_t
\end{equation}
where $\star$ is the correlation operation.
We follow the work of \cite{weston2022fastmbym}, which decouples the correlation operation for the rotation retrieval $R^*$ and the translation $t^*$ by exploiting the intrinsic translational invariance of the Fourier Transform.
This procedure has huge advantages in run-time performances and memory efficiency compared to the original work \cite{barnes2020masking} which performed an exhaustive brute-force search.

Importantly, the system is based on the assumption that $Z_t \approx T_{[R|t]}(Z_{t-1})$; yet, this is challenged by the very nature of radar, where noise, occlusions or multi-path reflections can make the two scans, although taken close spatially, very different visually.
To mitigate such an issue, \cite{barnes2020masking,weston2022fastmbym} introduce a learnt masking process to eliminate portions of the scans that would make the motion estimation inaccurate.

Let $G$ be an encoder-decoder network that returns a mask scoring each pixel of a radar scan with a scalar in the range $[0, 1]$.
The masking network can be applied to $Z_{t-1}$ and $Z_t$ to retrieve the relative masks $M_{t-1}$ and $M_t$, which can then be multiplied element-wise to the raw scans to retrieve $S_{t-1}$ and $S_t$; these latter can be used then in the scan-matching pipeline to retrieve $[R | t]$ from \cref{eq:correlatoin}.
Importantly, as the scan-matching pipeline is completely differentiable, $G$ can be learnt from an odometry signal through $L_1$ loss.

Whereas the original pipeline used a concatenation of $Z_{t-1}$ and $Z_t$ as input to $G$, we propose adopting a Siamese network architecture to inform the network through Doppler information; this pipeline can be seen in \cref{fig:2_channel_diagram}.
Specifically, $G$ receives $Z_{t-1}$ and $Z_t$ singularly to create $M_{t-1}$ and $M_t$ as 2-channel images, where the first channel contains the 200 azimuths modulated with pattern 1 and the second with pattern 2 (see \cref{fig:modulations}).
The resulting masks are then applied to the first channel to create $S_{t-1}$ and $S_t$.
This approach allows the network to exploit Doppler information implicitly from the raw signals while maintaining the benefits from a fully-differentiable approach and consistency in terms of pattern modulation when cross-correlating the radar images.

\subsection{Doppler-Aware Velocity Estimation (DAVE-NN)}
We propose to extract motion estimation directly from one radar scan containing Doppler information.
Whereas the previous experiments relied on geometry matching, this approach regresses the pose $[R | t]$ end-to-end from a \textit{single} 2-channel Doppler radar scan, which naturally exposes the network to the ``zig-zag'' patterns essential to extracting odometry information.

Our Doppler-aware Velocity Estimation network (DAVE-NN) is directly inspired from \cite{ruiz2018fine}, where the authors use an intermediate classification step to regress the head orientations of users.
In the same way, we trained our network to learn odometry by using an $L1$ loss on the softmax-weighted average of the classification results.
Differently than \cite{ruiz2018fine}, we did not apply a classification loss as the system did not empirically achieve better results.

Notably, we decided to include all three components of the motion estimation $t_x$, $t_y$ and $\theta$ to test the ability of the network to retrieve them from Doppler.
Physically, only translation can be estimated from it, as the velocity shift caused by Doppler happens only in the radial direction.
Nevertheless, we included the rotation to test if other effects, such as the first-last azimuth seam seen in \cref{fig:doppler_normal_val_scan}, can lead to an orientation estimation.

\subsection{Estimation fusion}

To combine the benefits of scan-matching and direct regression using Doppler information, we also propose an approach that fuses both methods.
This is done using a complementary filter approach, where confidences are extracted from both methods by considering the sharpness of the pose estimate, which is then used to weight the translation estimate from each method.
We limit ourselves in the forward dimension of motion for the estimate fusion, although the approach can be extended to the lateral direction as well.

The different estimation procedures encode their confidence in the longitudinal pose estimate differently.
The scan-matching approach outputs a correlation volume $C_{xy} \in \mathbb{R}^{n \times n}$ over possible translations ($t_x$ and $t_y$).
DAVE-NN outputs a logits vector $l_x \in \mathbb{R}^{b}$, containing classification scores for all $b$ possible values of $t_x$.

Firstly, the correlation volumes are reduced over the $y$-dimension  with a 
$\max$ operation, followed by a bilinear interpolation to the same dimensionality as the logits, resulting in $C_x \in \mathbb{R}^{b}$.
Correlation and logit values are distributed differently, so they are both normalised to have zero-mean and unit variance.
The confidence values for each are calculated as $c_S = \max [\sigma_{\tau_{S}}(C_x)]$ and $\ c_D = \max[\sigma_{\tau_D}(l_x)]$ for scan-matching and direct regression respectively, where $\sigma_{\tau}$ is the softmax function with a given temperature $\tau$.

In a complementary filter, the estimate weights must sum to 1; therefore, the weights are calculated as $[w_S, w_D] = \sigma_{\tau_{w}} ([c_S, c_D])$.
Finally, $t_x$ is calculated as follows from the scan-matching estimate $[t_x]_{S}$ and direct regression estimate $[t_x]_{D}$:
\begin{equation}
    t_x = w_S [t_x]_{S} + w_D [t_x]_{D} 
\end{equation}

\section{Experimental setup}
In this section, we report details on the implementation and training regimes and our evaluation methodology.

\subsection{Implementation details}

Our approach is implemented using the PyTorch library, more specifically though MMSegmentation\footnote{\url{github.com/open-mmlab/mmsegmentation}} and TIMM\footnote{\url{github.com/rwightman/pytorch-image-models}}, which provide a collection of popular deep-learning modules.
We use an Adam optimiser for training both networks.

As backbones for the masking network $G$ and DAVE-NN, we employed EfficientNet-B0 \cite{tan2020efficientnet}.
$G$ then decodes the embedding using an \gls{aspp} head \cite{chen2017rethinking}, while DAVE-NN uses five linear layers of sizes $[512, 1024, 512, 256, 128]$ with ReLU activation followed by three separate heads for each variable, with sizes $[256, 127]$.
The output is a classification score, which corresponds to the values in the ranges $[-1, 5]$ \si{\metre} for $t_x$, $[-1, 1]$ \si{\metre} for $t_y$ and $[-0.5, 0.5]$ \si{\radian} for $\theta$.
For both networks, we input radar images of size $255 \times 255$.

For determining the temperature values it is observed that, in most scenes, scan-matching is typically more accurate, therefore the temperature values are weighted towards this: $\tau_{S}=1$, $\tau_{D}=1.2$.
In order to reject pose estimation failure, $\tau_w$ is set to 0.0001, thereby severely down-weighting unconfident estimates.

\subsection{Datasets}
In conjunction with \gls{rdd}, we use the Radar RobotCar Dataset \gls{rrcd} \cite{barnes2020oxford} to validate our masking pipeline and understand the challenging nature of \gls{rdd}.
For \gls{rrcd}, we employ the training and validation routes designed in \cite{tang2020rsl} and will report the results on the test split of the run \texttt{2019-01-10-12-32-52}.
Instead, results on \gls{rdd} are reported on the test split of the run \texttt{log\_13}, composed of a rural driving scene with moderate traffic.
This run includes a narrow tunnel of trees -- shown in \cref{fig:doppler_normal_test_scan} -- which challenges geometrical scan matches.

\subsection{Baselines and Metrics}

To evaluate our Doppler-aware scan-matching method, \texttt{2-channel-doppler}, and the fused system, \texttt{fused}, we compare them against four baselines:
\begin{enumerate}
    \item \texttt{raw}: this baseline does not apply masking to the radar scans.
    \item \texttt{no-doppler-400}: this baseline mimics the original work \cite{weston2022fastmbym} with the updated masking network to verify the goodness of the results.
    \item \texttt{no-doppler-200}: the same network is now applied to scans with a reduced azimuth number: for \gls{rrcd} we downsample the azimuth direction by two and for \gls{rdd} we extract only the azimuths with one modulation.
    The rationale behind this baseline is to understand how much Doppler information does benefit the process.
    \item \texttt{full-scan}: this baseline directly applies the masking network and cross-correlation process to the radar scan without splitting the two modalities into separate channels.
    This baseline shows how a more naive approach does not benefit the system.
\end{enumerate}

We use the odometry metrics proposed for the KITTI odometry benchmark \cite{geiger2012we}, i.e. rotational and translational errors, as the main indicators of the accuracy of the odometry algorithms.

\section{Experimental results}
Here, we discuss the results of this work.

\subsection{Reduced Azimuth number}

\Cref{tab:kitty_rrcd} shows the results of our pipeline applied to the \gls{rrcd} to validate our segmentation network.
Indeed, the original work employed a UNet, while we used a DeepLabv3 network \cite{chen2017rethinking}.
Our system reaches comparable odometry results with the original implementation, even when the number of azimuths is halved to resemble the \gls{rdd}.
On this note, we can see how the same network, trained and tested on the \gls{rdd}, reaches far worse results, highlighting how challenging the new dataset is for a geometric scan-matching process.
We blame two main factors: the reduced training dataset compared to the \gls{rrcd} and the feature-poor rural scenery, as exemplified in \cref{fig:doppler_normal_test_scan}.

\begin{table}
\centering
\renewcommand{\arraystretch}{1.3}
\begin{tabular}{l|c|c}
 Model & Rot. error $^\circ$/km & Trans. Error $\%$ \\
 \hline
Original results \cite{weston2022fastmbym} -- \gls{rrcd} & 6.30 & 2.00 \\
\texttt{raw} -- \gls{rrcd} & 17.46 & 2.42  \\
\texttt{no-doppler-400} -- \gls{rrcd} & 8.06 & 1.38  \\
\texttt{no-doppler-200} -- \gls{rrcd} & 5.56  & 0.91  \\
\texttt{raw} -- \gls{rdd} & 34.83  & 58.35  \\
\texttt{no-doppler-200} -- \gls{rdd} & 32.19  & 28.03 \\

\end{tabular}

\caption{KITTI metrics to validate our approach, evaluating the test sets of the \gls{rrcd} and \gls{rdd}. We achieve similar performances to the reported accuracy values of the original work \cite{weston2022fastmbym}, even with half the number of azimuths, to allow direct comparison to the \gls{rdd}. Here, we see stark worse results due to the fewer training data points and the more challenging scenery.\label{tab:kitty_rrcd}}
\end{table}

\begin{table}
\begin{center}
\title{KITTI Metrics for Full Scans on the \gls{rdd}.}
\renewcommand{\arraystretch}{1.3}
\begin{tabular}{l|c|c}
 Model & Rot. Error $^\circ$/km & Trans. Error \\
 \hline
 \texttt{raw} & 34.83  & 58.35  \\
 \texttt{no-doppler-200} & 32.19  & 28.03  \\
 \texttt{full-scan} & 35.96  & 19.20  \\
 \texttt{2-channel-doppler} & 35.83  & 5.98  \\
 \texttt{fused} & 35.83  & 4.59  \\
\end{tabular}
\caption{\label{table:kitti_rdd}KITTI metrics to evaluate the benefits of using Doppler information. We can see how the usage of Doppler can enhance the accuracy results, but only if used taking into consideration the nature of the scan-matching process.}
\end{center}
\vspace{-15pt}
\end{table}

\subsection{Doppler-aware masking network selection}
\begin{figure*}[t]
\centering
\begin{subfigure}{.5\textwidth}
  \centering
    \includegraphics[width=1.0\textwidth]{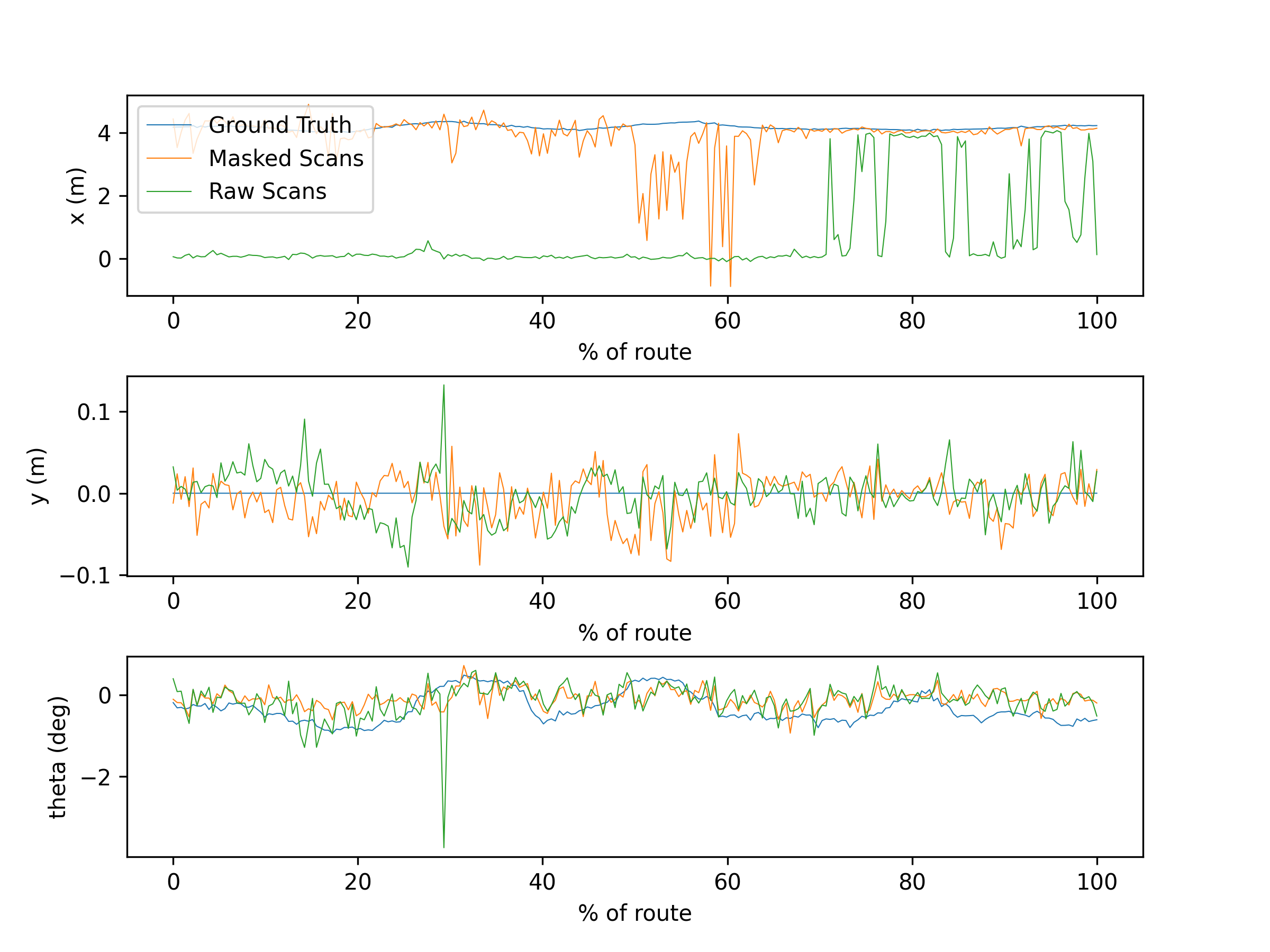}
  \caption{\label{fig:masking_plots}}
\end{subfigure}%
\begin{subfigure}{.5\textwidth}
  \centering
  \includegraphics[width=1.0\textwidth]{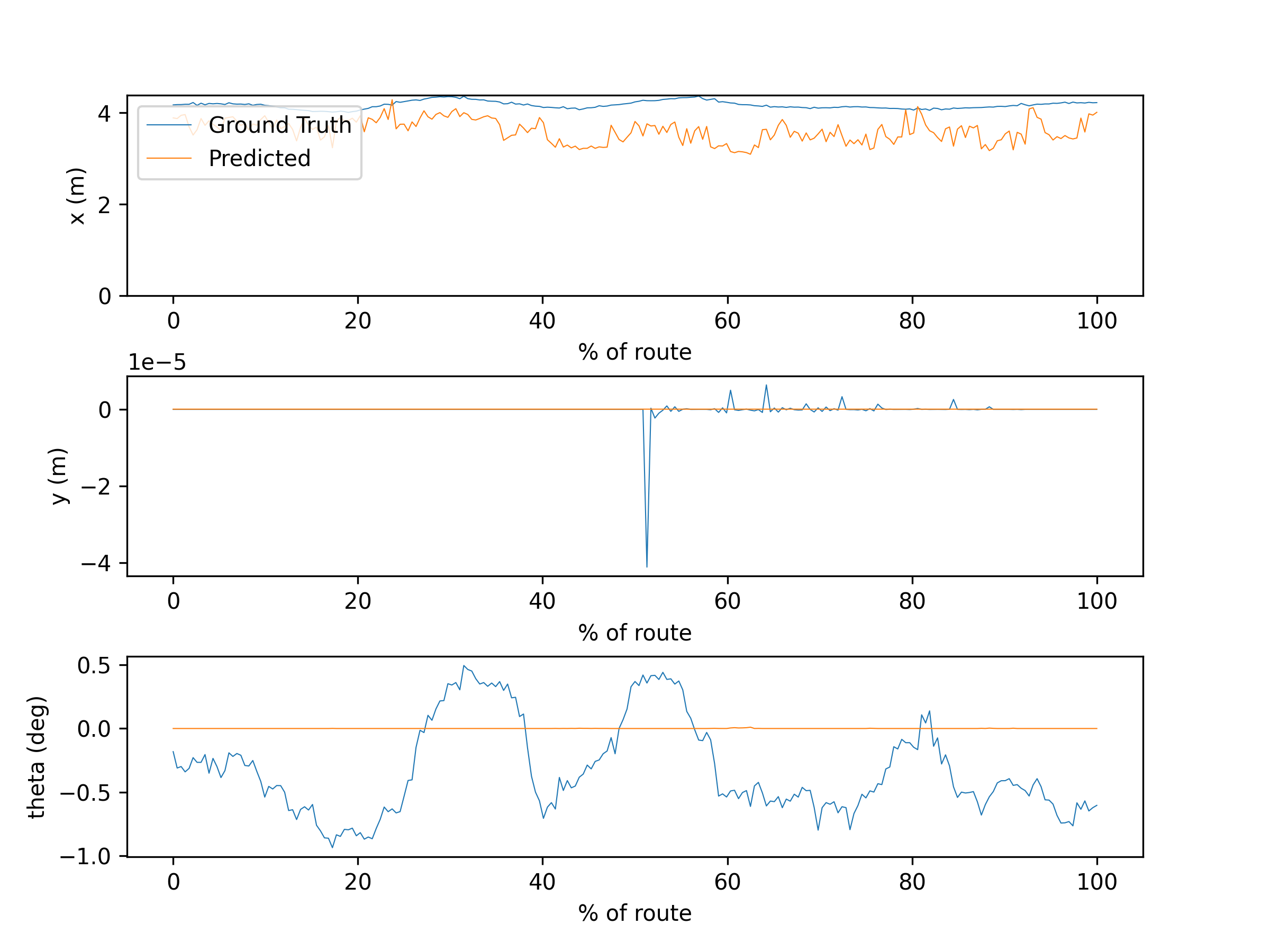}
  \caption{\label{fig:dave_plots}}
\end{subfigure}
\caption{Odometry signals for the test set of the \gls{rdd} for our (a) Doppler-aware masking network compared with using only raw scans and (b) DAVE-NN. The masking process is clearly improving the cross-correlation odometry estimation; yet, many failure points are present in the route (around 50\% through it). This is due to feature-poor scenery as in \cref{fig:doppler_normal_test_scan}. Instead, DAVE-NN recovers lower-quality odometry estimations in terms of accuracy but does not present any clear failure point. On the other hand, as expected, it cannot estimate angular displacements.\label{fig:odom_plots}}
\end{figure*}

\Cref{table:kitti_rdd} shows the results of using Doppler information during motion estimation.
We can see how using Doppler information naively does not bring any benefit, while our approach reduced the translation error three-fold.
The estimates from the proposed approach are shown in \cref{fig:masking_plots} against using the raw scans with no Doppler information.
The benefit is clear: the forward motion estimation using raw scans could not follow the ground truth for most of the run, while the proposed approach exhibits high accuracy.
Nevertheless, many failure points are present, due to the inability of the geometric approach to deal with ambiguous scenery, as the narrow tunnel depicted in \cref{fig:doppler_normal_test_scan}.

\subsection{Doppler-only motion estimation}

\Cref{fig:dave_plots} show the qualitative results for applying DAVE-NN to estimate odometry in the test set.
Firstly we notice how the forward motion estimation does follow the ground truth, although underestimating it slightly; nevertheless, most importantly, no catastrophic failures are present, as happens during scan  matching.
Instead, the angular estimation is not working; as expected, the Doppler information cannot assist in estimating the rotation and the other artefacts are not sufficient either.
For this reason, we do not report any metrics as they would be unfair to the system.

\subsection{Estimation fusion}

\begin{figure}
\centering
  \includegraphics[width=0.45\textwidth]{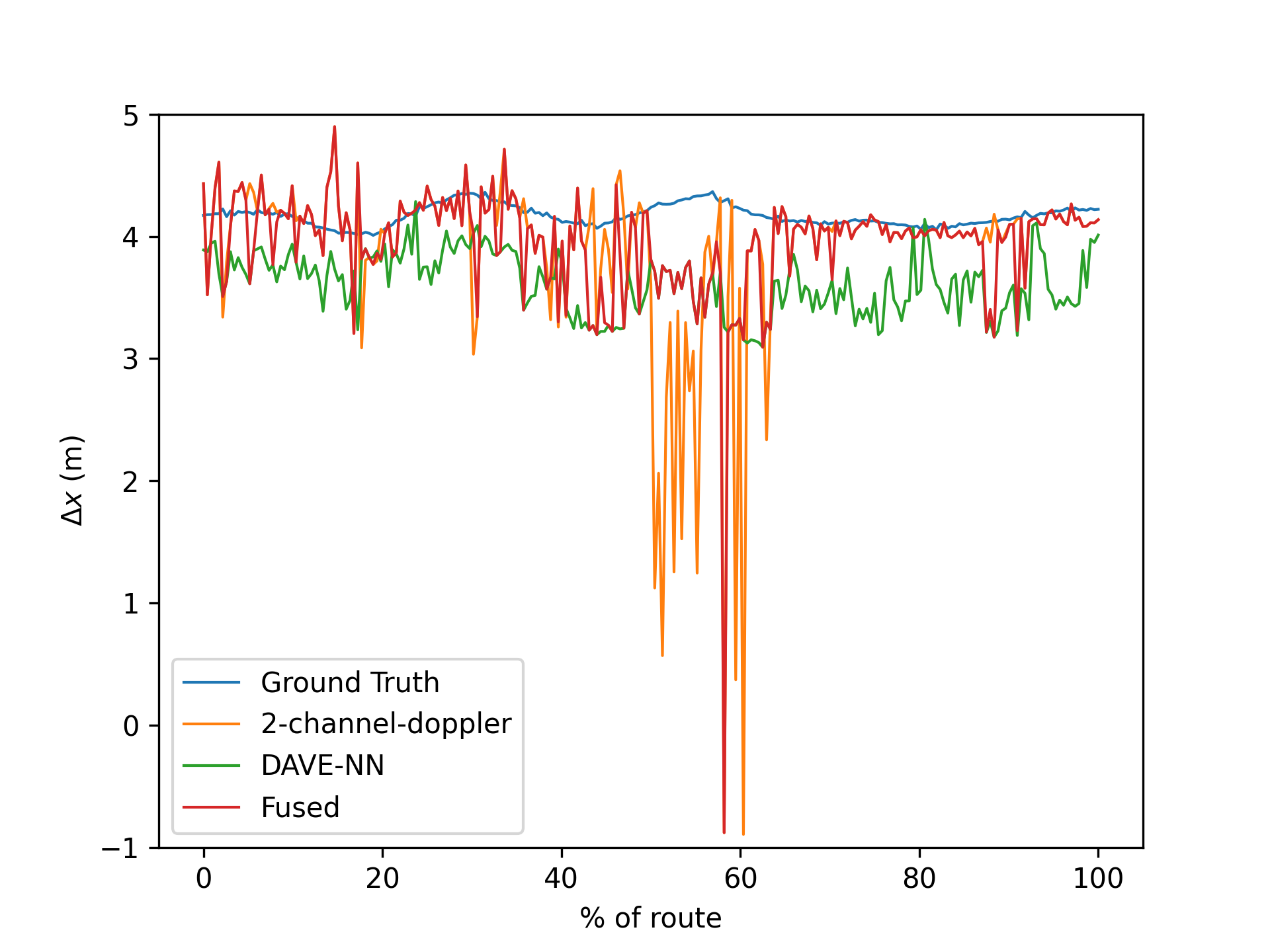}
  \caption{Odometry signal for the two separate systems and fused together. It is noticeable how all the catastrophic failures have been removed.\label{fig:fusion_plot}}
\end{figure}

Using an estimate of the uncertainty to fuse the odometry estimation for the forward motion improves the results in terms of failure mitigation.
Indeed, \cref{fig:fusion_plot} shows how all the catastrophic failures have been removed, while \cref{table:kitti_rdd} shows how the fused approach -- where only $f_x$ is fused -- has slightly lower accuracy than masking alone.
Indeed, DAVE-NN gives less accurate results overall, which, when integrated over time, lead to slightly worse metrics regardless of the improved robustness.

\section{Conclusions}
We studied the beneficial effect that Doppler information contained in \gls{fmcw} scanning radar scans can have on \gls{ro}.
Here, we propose to treat Doppler as an additional channel in helping a masking function clean radar data before applying correlative scan matching to recover the transformation between two scans.
Moreover, we propose a purely-regression approach to extract translational estimates from \textit{single} scans with Doppler information.
The first approach improved accuracy compared to the same input without Doppler information, whereas the second showed high robustness to challenging scenes, such as narrow tunnels.
We finally fuse these two pieces of information using intrinsic uncertainties to maximise robustness and accuracy.
Our methodology is validated on a custom dataset, \gls{rdd}, composed of \gls{fmcw} radar scans with Doppler information and odometry ground-truth signal, which is released alongside this work.

\bibliographystyle{ieeetr}
\bibliography{biblio}

\begin{thebibliography}{10}

\bibitem{burnett2023boreas}
K.~Burnett, D.~J. Yoon, Y.~Wu, A.~Z. Li, H.~Zhang, S.~Lu, J.~Qian, W.-K. Tseng,
  A.~Lambert, K.~Y. Leung, {\em et~al.}, ``Boreas: A multi-season autonomous
  driving dataset,'' {\em The International Journal of Robotics Research},
  2023.

\bibitem{9197298}
G.~Kim, Y.~S. Park, Y.~Cho, J.~Jeong, and A.~Kim, ``Mulran: Multimodal range
  dataset for urban place recognition,'' in {\em IEEE International Conference
  on Robotics and Automation (ICRA)}, 2020.

\bibitem{barnes2020oxford}
D.~Barnes, M.~Gadd, P.~Murcutt, P.~Newman, and I.~Posner, ``The oxford radar
  robotcar dataset: A radar extension to the oxford robotcar dataset,'' in {\em
  Proceedings of the IEEE International Conference on Robotics and Automation
  (ICRA)}, 2020.

\bibitem{sheeny2020radiate}
M.~Sheeny, E.~De~Pellegrin, S.~Mukherjee, A.~Ahrabian, S.~Wang, and A.~Wallace,
  ``Radiate: A radar dataset for automotive perception in bad weather,'' in
  {\em 2021 IEEE International Conference on Robotics and Automation (ICRA)},
  2021.

\bibitem{gadd2021contrastive}
M.~Gadd, D.~De~Martini, and P.~Newman, ``Contrastive learning for unsupervised
  radar place recognition,'' in {\em International Conference on Advanced
  Robotics (ICAR)}, 2021.

\bibitem{de2020kradar}
D.~De~Martini, M.~Gadd, and P.~Newman, ``Kradar++: Coarse-to-fine fmcw scanning
  radar localisation,'' {\em Sensors}, 2020.

\bibitem{suaftescu2020kidnapped}
{\c{S}}.~S{\u{a}}ftescu, M.~Gadd, D.~De~Martini, D.~Barnes, and P.~Newman,
  ``Kidnapped radar: Topological radar localisation using
  rotationally-invariant metric learning,'' in {\em IEEE International
  Conference on Robotics and Automation (ICRA)}, 2020.

\bibitem{kaul2020rss}
P.~Kaul, D.~De~Martini, M.~Gadd, and P.~Newman, ``Rss-net: Weakly-supervised
  multi-class semantic segmentation with fmcw radar,'' in {\em IEEE Intelligent
  Vehicles Symposium (IV)}, 2020.

\bibitem{8794014}
R.~Aldera, D.~D. Martini, M.~Gadd, and P.~Newman, ``Fast radar motion
  estimation with a learnt focus of attention using weak supervision,'' in {\em
  International Conference on Robotics and Automation (ICRA)}, 2019.

\bibitem{broome2020road}
M.~Broome, M.~Gadd, D.~De~Martini, and P.~Newman, ``On the road: Route proposal
  from radar self-supervised by fuzzy lidar traversability,'' {\em AI}, vol.~1,
  no.~4, pp.~558--585, 2020.

\bibitem{williams2020keep}
D.~Williams, D.~De~Martini, M.~Gadd, L.~Marchegiani, and P.~Newman, ``Keep off
  the grass: Permissible driving routes from radar with weak audio
  supervision,'' in {\em IEEE International Conference on Intelligent
  Transportation Systems (ITSC)}, 2020.

\bibitem{tang2021self}
T.~Y. Tang, D.~De~Martini, S.~Wu, and P.~Newman, ``Self-supervised learning for
  using overhead imagery as maps in outdoor range sensor localization,'' {\em
  The International Journal of Robotics Research}, 2021.

\bibitem{tangauro2023}
T.~Y. Tang, D.~De~Martini, and P.~Newman, ``Point-based metric and topological
  localisation between lidar and overhead imagery,'' {\em Autonomous Robots
  (AURO)}, 2023.

\bibitem{8460687}
S.~H. Cen and P.~Newman, ``Precise ego-motion estimation with millimeter-wave
  radar under diverse and challenging conditions,'' in {\em IEEE International
  Conference on Robotics and Automation (ICRA)}, 2018.

\bibitem{cen2019radar}
S.~H. Cen and P.~Newman, ``Radar-only ego-motion estimation in difficult
  settings via graph matching,'' in {\em International Conference on Robotics
  and Automation (ICRA)}, 2019.

\bibitem{9100374}
M.~Adams, E.~Jose, and B.-N. Vo, {\em Robotic Navigation and Mapping with
  Radar}.
\newblock 2012.

\bibitem{alhashimi2021bfarbounded}
A.~Alhashimi, D.~Adolfsson, M.~Magnusson, H.~Andreasson, and A.~J. Lilienthal,
  ``Bfar-bounded false alarm rate detector for improved radar odometry
  estimation,'' {\em arXiv preprint arXiv:2109.09669}, 2021.

\bibitem{cfear}
D.~Adolfsson, M.~Magnusson, A.~W. Alhashimi, A.~J. Lilienthal, and
  H.~Andreasson, ``{CFEAR} radarodometry - conservative filtering for efficient
  and accurate radar odometry,'' in {\em Proceedings of the IEEE/RSJ
  International Conference on Intelligent Robots and Systems}, 2021.

\bibitem{barnes2020under}
D.~Barnes and I.~Posner, ``Under the radar: Learning to predict robust
  keypoints for odometry estimation and metric localisation in radar,'' in {\em
  IEEE international conference on robotics and automation (ICRA)}, 2020.

\bibitem{burnett_radar_odom_2021}
K.~Burnett, D.~J. Yoon, A.~P. Schoellig, and T.~D. Barfoot, ``Radar odometry
  combining probabilistic estimation and unsupervised feature learning,'' in
  {\em Robotics: Science and Systems (RSS)}, 2021.

\bibitem{barnes2020masking}
D.~Barnes, R.~Weston, and I.~Posner, ``Masking by moving: Learning
  distraction-free radar odometry from pose information,'' in {\em {C}onference
  on {R}obot {L}earning ({CoRL})}, 2019.

\bibitem{weston2022fastmbym}
R.~Weston, M.~Gadd, D.~D. Martini, P.~Newman, and I.~Posner, ``Fast-mbym:
  Leveraging translational invariance of the fourier transform for efficient
  and accurate radar odometry,'' in {\em IEEE International Conference on
  Robotics and Automation (ICRA)}, 2022.

\bibitem{9327473}
K.~Burnett, A.~P. Schoellig, and T.~D. Barfoot, ``Do we need to compensate for
  motion distortion and doppler effects in spinning radar navigation?,'' {\em
  IEEE Robotics and Automation Letters}, 2021.

\bibitem{zhuang20234d}
Y.~Zhuang, B.~Wang, J.~Huai, and M.~Li, ``4d iriom: 4d imaging radar inertial
  odometry and mapping,'' {\em IEEE Robotics and Automation Letters}, 2023.

\bibitem{wu2022picking}
Y.~Wu, D.~J. Yoon, K.~Burnett, S.~Kammel, Y.~Chen, H.~Vhavle, and T.~D.
  Barfoot, ``Picking up speed: Continuous-time lidar-only odometry using
  doppler velocity measurements,'' {\em IEEE Robotics and Automation Letters},
  2022.

\bibitem{ruiz2018fine}
N.~Ruiz, E.~Chong, and J.~M. Rehg, ``Fine-grained head pose estimation without
  keypoints,'' in {\em Proceedings of the IEEE conference on computer vision
  and pattern recognition workshops}, 2018.

\bibitem{tan2020efficientnet}
M.~Tan and Q.~Le, ``Efficientnet: Rethinking model scaling for convolutional
  neural networks,'' in {\em International conference on machine learning},
  2019.

\bibitem{chen2017rethinking}
L.-C. Chen, G.~Papandreou, F.~Schroff, and H.~Adam, ``Rethinking atrous
  convolution for semantic image segmentation,'' {\em arXiv preprint
  arXiv:1706.05587}, 2017.

\bibitem{tang2020rsl}
T.~Y. Tang, D.~De~Martini, D.~Barnes, and P.~Newman, ``Rsl-net: Localising in
  satellite images from a radar on the ground,'' {\em IEEE Robotics and
  Automation Letters}, 2020.

\bibitem{geiger2012we}
A.~Geiger, P.~Lenz, and R.~Urtasun, ``Are we ready for autonomous driving? the
  kitti vision benchmark suite,'' in {\em IEEE conference on computer vision
  and pattern recognition}, 2012.

\end{thebibliography}

\end{document}